\documentclass{article}
\usepackage{ijcai25}

\usepackage{times}
\usepackage{soul}
\usepackage{url}
\usepackage[hidelinks]{hyperref}
\usepackage[utf8]{inputenc}
\usepackage[small]{caption}
\usepackage{graphicx}
\usepackage{amsmath}
\usepackage{amsthm}
\usepackage{booktabs}
\usepackage{algorithm}
\usepackage{algorithmic}
\usepackage[switch]{lineno}

\usepackage[frozencache,cachedir=.]{minted}
\usepackage{amsmath}
\usepackage{amsfonts} 
\usepackage{subcaption}
\usepackage{multirow}
\usepackage{booktabs}

\title{\textsc{Balans:} Multi-Armed Bandits-based Adaptive Large Neighborhood Search for Mixed-Integer Programming Problems}

\author{
Junyang Cai$^1$\and
Serdar Kad{\i}o\u{g}lu$^{2,3}$
\and
Bistra Dilkina$^1$
\\
\affiliations
$^1$Department of Computer Science, University of Southern California\\
$^2$AI Center of Excellence, Fidelity Investments\\
$^3$Department of Computer Science, Brown University\\
\emails
\{caijunya, dilkina\}@usc.edu,
serdark@cs.brown.edu
}
\begin{document}

\maketitle

\begin{abstract}
Mixed-integer programming (MIP) is a powerful paradigm for modeling and solving various important combinatorial optimization problems. Recently, learning-based approaches have shown a potential to speed up MIP solving via offline training that then guides important design decisions during the search. However, a significant drawback of these methods is their heavy reliance on offline training, which requires collecting training datasets and computationally costly training epochs yet offering only limited generalization to unseen (larger) instances. In this paper, we propose \textsc{Balans}, an adaptive meta-solver for MIPs with online learning capability that does not require any supervision or apriori training. At its core, \textsc{Balans} is based on adaptive large-neighborhood search, operating on top of an MIP solver by successive applications of destroy and repair neighborhood operators. During the search, the selection among different neighborhood definitions is guided on the fly for the instance at hand via multi-armed bandit algorithms. Our extensive experiments on hard optimization instances show that \textsc{Balans} offers significant performance gains over the default MIP solver, is better than committing to any single best neighborhood, and improves over the state-of-the-art large-neighborhood search for MIPs. Finally, we release \textsc{Balans} as a highly configurable, MIP solver agnostic, open-source software. 
\end{abstract}

\section{Introduction}

Mixed-integer programming (MIP) can model many important combinatorial optimization problems, and hence, improving the efficiency of MIP solving is of great practical and theoretical interest. Broadly, solving MIPs falls under two main classes of algorithms: exact and meta-heuristic methods. For exact methods, branch-and-bound~\cite{land2010automatic}, and its extensions, are at the core of solving MIPs to optimality. When proving optimality is beyond reach, meta-heuristics, such as Large Neighborhood Search (LNS), offer an attractive alternative for finding good solutions and increasing scalability.

Incorporating learning-based methods into MIP solving has shown great potential to improve exact and heuristic methods. For exact branch-and-bound based methods, these include learning to branch~\cite{BENGIO2021405,khalil2016learning,lodi2017learning,DBLP:conf/aaai/KadiogluMS12,cai2024learning}, 
% 
%,DBLP:conf/aaai/KadiogluMS12,dash
% learning cut generation~\cite{tang2020reinforcement} and
and node selection~\cite{NIPS2014_757f843a}. For meta-heuristics, these include predicting and searching for 
%khalil2022mip,
solutions~\cite{ding2020accelerating,cai2024multi}, integrating machine learning into LNS on top of MIP solvers~\cite{song2020general,huang2023searching}, and learning to schedule heuristics within the branch-and-bound search 
tree~\cite{khalil2017learning,chmiela2021learning,hendel2022adaptive}.

The significant drawback of these learning-guided methods is their heavy dependency on offline training. Training is computationally costly, depends on carefully curating training datasets with desired properties and distributions, and has limited generalization. Moreover, training might even depend on using exact solvers in the first place to create the supervised datasets, which defeats the purpose of improving solving for very hard instances. Adapting offline learning-based methods to new distributions and domains remains a challenge, and hence, on-the-fly or online learning approaches to MIP solving are a much-needed alternative. This is exactly what we study in this paper. 

In this paper, we focus on solving MIPs via online adaptive methods that do not depend on any offline training. Recent research has shown the potential of meta-solvers that embed a MIP solver within LNS~\cite{song2020general,tong2024optimization} and consequently achieve better performance than default MIP via LNS(MIP). The state-of-the-art in this line of research is based on a single neighborhood definition, referred to as local branching relaxation~\cite{huang2023local}. However, the previous work showing that offline learning can improve LNS 
%nair2020solving,
\cite{song2020general,huang2023searching} suffers from the above-mentioned limitations. We take the idea of machine learning-guided LNS for MIP further and propose a meta-solver based on \textit{Adaptive} LNS (ALNS) operating on a MIP solver, i.e., ALNS(MIP), composed of a diverse set of neighborhood definitions that are driven by multi-armed bandits online learning policy. 

\begin{figure*}[t]
    \centering
    \includegraphics[width=\linewidth]{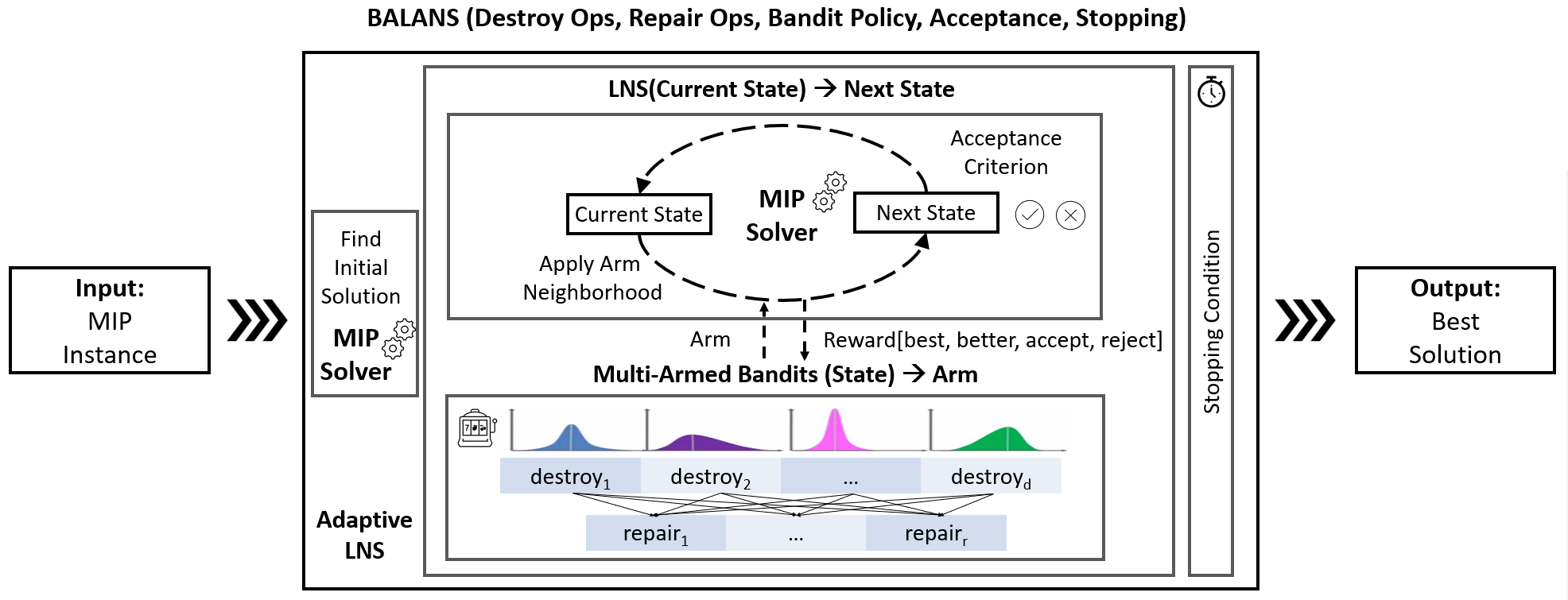}
% \vspace{-0.3cm}
\caption{The high-level architecture of the \textsc{Balans} Solver with its configurable components as input parameters.}
% (Destroy Operators, Repair Operators, Bandit Policy, Acceptance Criterion, Stopping Condition)}
\label{fig:balans}
% \vspace{-0.4cm}
\end{figure*}

% Removing per Bistra (too detailed in the intro)
% This is, however, non-trivial since we must now simultaneously decide which neighborhood to employ at every iteration among a large set of choices \textit{and} decide how to explore the search space by accepting or rejecting the proposed next move. These additional choices make the learning problem harder, but at the same time, given the significant drawbacks of offline learning, it must be tackled on the fly, adaptively for the specific instance at hand. 
This is, however, non-trivial since online learning is hard, but given the significant drawbacks of offline learning, it must be tackled on the fly, adaptively for the specific instance at hand. 
We show how to cast this as a \textit{multi-armed bandit problem} that treats adaptive neighborhoods as different arm choices with unknown reward distributions to be estimated during the search. Bringing these components together, we propose \textsc{Balans}, an online meta-solver for MIPs using multi-armed \textbf{B}andits-based \textbf{A}daptive \textbf{La}rge \textbf{N}eighborhood \textbf{S}earch operating across a diverse set of neighborhoods. 

Our main findings and contributions are as follows: 

\begin{itemize}
\item We show that the performance of our bandit-based ALNS(MIP), carefully implemented in our \textsc{Balans} solver, significantly improves the default MIP solver SCIP~\cite{bolusani2024scip}, outperforms LNS(MIP) that commits to any single neighborhood, and improves over the previous state-of-the-art LNS(MIP) approach on hard instances of optimization problems from different application domains.
\item We show that our bandit-based ALNS(MIP) rarely depends on the single best neighborhood and instead improves the performance by exploring and sequencing other weaker neighborhoods. 
% \item \bistra{Moreover, we show that the performance of our online approach matches, and in parts exceeds, the performance of offline methods that are specifically trained with supervised learning on the same distribution of instances. }
\item We perform ablation studies to show i) \textsc{Balans} is solver agnostic by performing the same set of experiments on a different MIP solver Gurobi~\cite{gurobi}, and ii), \textsc{Balans} is highly different from scheduling heuristics within the BnB tree~\cite{hendel2022adaptive}.
\item Finally, we release \textsc{Balans}\footnote{https://github.com/skadio/balans} as an open-source meta-solver for conducting ALNS(MIP) available to other researchers and practitioners with a one-liner from PyPI\footnote{pip install balans}.
\end{itemize}

More broadly, our \textsc{Balans} solver subsumes the previous literature on LNS(MIP) when run with a single neighborhood while serving as a highly configurable, modular, and extensible integration technology at the intersection of adaptive search, meta-heuristics, multi-armed bandits, and mixed-integer programming. 

In the following, let us start with a brief background and present our approach, followed by experimental results, discussions, and related work.

\section{Background}

% We briefly cover the necessary background on mixed-integer programming, adaptive large-neighborhood search, and multi-armed bandits. 

% \subsection{Mixed-Integer Programming (MIP)}
% \noindent \textbf{Mixed-Integer Programming (MIP):}
\paragraph{Mixed-Integer Programming (MIP):}
We deal with MIP solving that formulates optimization problems of the form:
% \vspace{-0.15cm}
\begin{align}
    f(x) = \min\{c^Tx\mid Ax\leq b, x\in \mathbb{R}^n, x_j \in \mathbb{Z} \ \forall j \in I \}\nonumber
\end{align}
where $f(x)$ is the objective function, and $A \in \mathbb{R}^{m\times n}, b \in \mathbb{R}^m, c \in \mathbb{R}^n$, and the non-empty set $I \subseteq {1, ..., n}$ indexes the integer variables.

% When $V=I$, we deal with Integer Programs (IPs), and when $V=B$, we deal with Binary Programs (BP). 
The Linear Programming (LP) relaxation of a MIP $x_{lp}$ is obtained by relaxing integer variables to continuous variables, i.e., by replacing the integer constraint $x_j \in \mathbb{Z} \ \forall j \in I$ to 
$x_j \in \mathbb{R} \ \forall j \in I$. The LP relaxation is an essential part of the branch-and-bound (BnB) algorithm, and many destroy operators are used in large-neighborhood searches.

% \subsection{(Adaptive) Large Neighborhood Search (ALNS)}

% \noindent \textbf{(Adaptive) Large-Neighborhood Search (LNS / ALNS):}
\paragraph{(Adaptive) Large-Neighborhood Search (LNS / ALNS):}
LNS is a meta-heuristic that starts with an initial solution and then iteratively destroys and repairs a part of the solution until hitting a stopping condition~\cite{pisinger2019large}. When applied to MIPs, LNS(MIP), we first generate an initial solution, typically found by running BnB for a short time. This becomes the starting point of LNS. In iteration $t\geq1$, given the solution of the previous state, $x_{t-1}$, a sub-MIP is created by a destroy heuristic. The re-optimized solution to this sub-MIP, $x_t$, then becomes the candidate for the next state, decided by an accepted criterion. 
% If accept, it becomes the current state solution $x_t$. If not, $x_t = x_{t-1}$. 
Adaptive LNS (ALNS) is defined by multiple destroy heuristics to choose from at each iteration, thereby extending LNS(MIP) to ALNS(MIP).
% On difficult optimization instances, when proving optimality remains out of reach, LNS is more effective than BnB in improving the objective~\cite{song2020general}.

% \subsection{Multi-Armed Bandits (MAB)}
% \noindent \textbf{Multi-Armed Bandits (MAB):}
\paragraph{Multi-Armed Bandits (MAB):}
MAB algorithms~\cite{vermorel2005multi} is a class of reinforcement learning aimed at solving online sequential decision-making problems. In MAB, each \textit{arm} defines a decision that an agent can make, generating either a deterministic or stochastic \textit{reward}. At each step, the agent faces a decision whether to utilize an arm with the highest expected reward (``exploit”) or to try out new arms to learn something new (``explore”). The agent’s goal is to maximize the cumulative reward in the long run, for which it must balance exploration and exploitation. The reward of each arm is estimated from past decisions using a learning policy~\cite{DBLP:journals/corr/KuleshovP14}. 
% Several learning policies such as $e$-Greedy, Softmax, Thompson Sampling are designed to estimate the reward of arms based on previous decisions.

\section{\textsc{Balans}: Online Meta-Solver for MIPs}

As illustrated in Figure~\ref{fig:balans}, our \textsc{Balans} approach brings the complementary strengths of MIP, ALNS, and MAB together to create an online meta-solver for tackling hard combinatorial optimization problems.

Given an MIP instance as input, we first find an initial solution by running an MIP solver with a time limit to find the first feasible solution. This also allows filtering easy-to-solve instances up front. As part of this process, we instantiate a \textit{single} underlying MIP model that we maintain throughout the entire search. Similarly, the LP relaxation at the root node is saved together with other information, such as the indexes of binary, discrete, and continuous variables. This incremental nature of \textsc{Balans} solver is important for good performance, as shown later in our experiments. 

The initial solution yields the current search state to start operating ALNS. As depicted in Figure~\ref{fig:balans}, ALNS is a combination of LNS guided by MAB to adapt to diverse neighborhoods. For this purpose, we designed a diverse set of destroy operators and a single repair operator. The repair operator uses the MIP solver to re-optimize the instance resulting from the destroy operation. Next, we overview destroy operators. 

% All our operators are implemented in incremental fashion, unlike the previous work~\cite{huang2023local, huang2023searching}. During the whole ALNS pipeline, we only keep one single copy of the instance and we are not reloading the model every time we destroy the problem. Instead, we keep track of the modification and undo the modification after the destroy operation. With some empirical testing, we observed two fold speedup during ALNS iterations, especially on larger instances.

\begin{figure*}[ht]
    \centering
    \includegraphics[width=\linewidth]{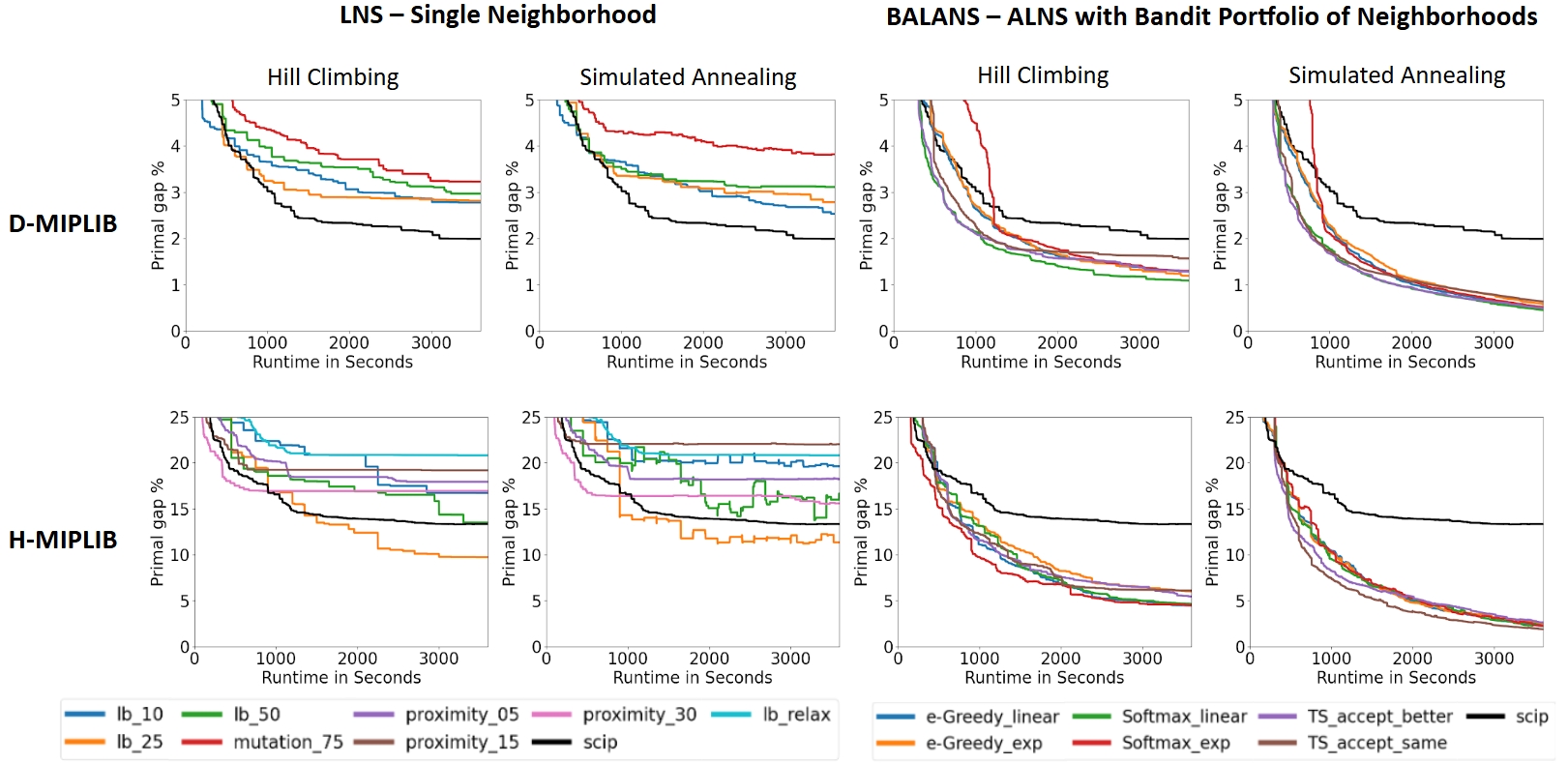}
\caption{The primal gap (the lower, the better) as a function of time, averaged over instances from D-MIPLIB (top row) and H-MIPLIB (bottom row). We compare LNS (four plots on the left) that commits to a single neighborhood vs. \textsc{Balans} (four plots on the right) that uses ALNS with a bandit portfolio of neighborhoods. For LNS, each line represents a different neighborhood. For \textsc{Balans}, each line represents a different learning policy. Both approaches are run with Hill Climbing and Simulated Annealing. Baseline, LNS and \textsc{Balans} use SCIP as the base MIP solver. Within each dataset (row), the SCIP baseline, shown as the black line, is identical.}
\label{fig:lines}
% \vspace{-0.3cm}
\end{figure*}
\subsection{Destroy Operators}
% As listed in Table~\ref{tab:destroy}, 
We design eight built-in destroy operators in \textsc{Balans}. These include Crossover~\cite{rothberg2007evolutionary}, DINS~\cite{ghosh2007dins}, Local Branching~\cite{fischetti2003local}, Mutation~\cite{rothberg2007evolutionary}, Proximity Search~\cite{fischetti2014proximity}, Random Objective, RENS~\cite{berthold2014rens}, and RINS~\cite{danna2005exploring}. 

Conceptually, destroying neighborhoods is aimed at creating a sub-MIP that is easier than the original problem. This is achieved by either altering the objective function, as done in Proximity Search and Random Objective, or by adding constraints based on a certain set of destroy variables, as done in other operators. We adapted the original definitions of these operators for the ALNS(MIP) context. The main idea behind our adaptations is to ensure that the operator does not attempt an identical neighborhood in subsequent iterations. For example, local branching is inherently deterministic, and we modify it by introducing a randomized destroy size in every iteration.
% We also introduce a configurable parameter in most of operators to allow different size of neighborhood definition. 
We provide the precise definitions and details of the selection of destroy heuristics in Appendix B~\cite{cai2024balans}. 
% of our destroy operators used in \textsc{Balans}. 
% and we provide more details in the supplementary material.

Let us note that there are destroy neighborhood operators that are not implemented in \textsc{Balans}. Most notable is the Local Branching Relaxation (lb-relax) from~\cite{huang2023local}. This is the state-of-the-art LNS(MIP) approach. 
%It mostly resembles Local Branching but also incorporates elements from DINS, which leverages the LP relaxation and tries to bound the solution of the next iteration around the LP solution, and Mutation, which introduces a randomized selection among discrete indices. 
However, lb-relax has a hyper-parameter to control the destroy size that must be chosen carefully for each problem domain. In addition, the initial destroy size is then dynamically adjusted during the search according to a fixed schedule. The hybrid nature of lb\_relax combined with a tuned destroy size and its dynamic adjustment is key to its state-of-the-art LNS(MIP) performance. 

In contrast, in \textsc{Balans}, our destroy operators are not mixing different flavors together and are designed to be distinct to constitute a diverse portfolio. In principle, an effective online learning algorithm would be able to sequence these distinct operators in a way to obtain the desired hybrid behavior for the instance at hand. Some destroy operators such as DINS, Local Branching, and Proximity work only on specific MIP subfamilies, e.g. binary or integer only. When combined together in \textsc{Balans}, we cover all subfamilies of MIPs. Moreover, notice that we do not need to tune for the destroy size, a common requirement of previous work. We simply introduce the same operator multiple times in our portfolio with varying destroy sizes, serving as different options to choose from during the search. 
% Overall, from these eight unique destroy neighborhoods, we obtain sixteen different destroy operators by varying their parameters as listed in experiments section.
The question then becomes how to learn effectively during search and choose among different options on the fly.
% , which we discuss next.
%, as discussed next. 

% their Adaptive in ALNS means they introduce some parameters to control the neighborhood size and the parameter is adaptive changing due to the performance in every LNS iteration. This introduce a lot more hyper-parameter into the LNS and these parameters needs to be carefully chosen for each distribution of the problem. We want to create a general MIP meta-solver can applied to any instances, so we did not introduce this adaptive idea into our implementation. Instead, we introduce a configurable parameter in most of operators to allow different size of neighborhood definition and we directly put all the different configuration destroy operators to our \textsc{Balans}.

\subsection{Online Learning}
We identify two types of exploration that are crucial for effective online learning. The first exploration decision stems from state exploration, which decides whether the search should continue with the next state or discard the move. The second exploration decision stems from neighborhood exploration, which decides the destroy neighborhood operator to apply at each state. Distinguishing the two different needs of exploration and addressing them separately in a principled manner is a novelty of our approach and key to good performance.

% \smallskip
% \noindent \textbf{State Exploration: }
\paragraph{State Exploration:}
 As shown in Figure~\ref{fig:balans}, the acceptance criterion governs the search space among states. We consider two complementary acceptance criteria: Hill Climbing (HC) and Simulated Annealing (SA). HC mostly exploits yet allows the search to progress to the next state when the objective value is the same. On the other hand, SA offers more exploration capacity and allows the search to move to worsening next states.  

% \smallskip
% \noindent \textbf{Neighborhood Exploration:} 
\paragraph{Neighborhood Exploration:}
We employ multi-armed bandits for choosing among different neighborhoods. To apply MAB, there are three important design decisions: the definition of \textit{arms}, the \textit{reward mechanism}, and the \textit{learning policy}.

As shown in Figure~\ref{fig:balans}, in \textsc{Balans}, we treat \textit{every pair} of destroy and repair operators as a single arm. When an arm is selected, the neighborhood operation is \textit{atomic}; it destroys the current state and then immediately repairs it to obtain the next state. In this paper, given that we only have a single repair operator, which re-optimizes the sub-MIP obtained from applying to destroy, the list of arms can be treated identically with our destroy operators. If desired, users of \textsc{Balans} can introduce additional repair operators.

Our reward mechanism for MAB is designed specifically for the ALNS(MIP) framework. 

We consider four distinct rewards aligned to the possible outcomes of the acceptance criterion as part of state exploration. These four distinct outcomes are: the next state is the best solution found so far, the next state is better than the previous state, the next state is neither the best nor better, but it is accepted as a move, or the next state is rejected. Each outcome is associated with a specific reward value, as later listed in our experiments. Finally, MAB learns from historical arm choices associated with their observed rewards based on its learning policy. 

%For learning policy, we consider three strategies: e-Greedy~\cite{auer2002finite} and Softmax~\cite{luce1977choice} with numeric rewards and Thompson Sampling~\cite{thompson1933likelihood} with binary rewards. For numeric rewards, we experiment with linear and exponential settings. 

To summarize our overall algorithm, once the initial solution is found, the main ALNS loop starts, which is an interplay between LNS and MAB. First, MAB selects an arm among the different neighborhoods and provides it to LNS. Then, LNS applies the operation and observes the outcome. Based on the resulting state, its solution quality, and the acceptance criterion, LNS provides the reward feedback to MAB. MAB updates the reward estimate of the selected arm according to its learning policy. This process repeats until hitting the stopping condition, and at the end, the best solution found during the search is returned. 

\begin{table*}[t]
\centering
\renewcommand{\arraystretch}{1.2}
\begin{tabular}{@{}ccccccccc@{}}
\toprule
                     & \multicolumn{4}{c}{Distributional MIPLIB (D-MIPLIB)}                        & \multicolumn{4}{c}{MIPLIB Hard (H-MIPLIB)}                                  \\ \cmidrule(l){2-9} 
                     & \multicolumn{2}{c}{Hill Climbing} & \multicolumn{2}{c}{Simulated Annealing} & \multicolumn{2}{c}{Hill Climbing} & \multicolumn{2}{c}{Simulated Annealing} \\ \cmidrule(l){2-9} 
                     & PG (\%) ↓         & PI ↓            & PG (\%) ↓        & PI ↓                 & PG (\%) ↓          & PI ↓             & PG (\%)↓             & PI ↓               \\ \midrule
SCIP                 & 2.11±1.68        & 105±56.5       & N/A                & N/A                & 13.3±3.95        & 514±134        & N/A                 & N/A               \\
lb\_relax            & 5.72± 3.25       & 163±73.6       & N/A                & N/A                & 24.2±15.5        & 778±156        & N/A                 & N/A               \\ \midrule
local\_branching\_25 & 2.80±2.27        & 110±81.2       & 2.77±2.03          & 115±77.4           & 9.73±2.11        & 485±95.9       & 11.4±2.34           & 502±93.0          \\
mutation\_75         & 2.21±3.98        & 104±166        & 2.81±4.9           & 114±176            & 57.7±139         & 971±473        & 58.8±139            & 991±472           \\
proximity\_15         & 29.6±27.8        & 771±253        & 29.6±27.8          & 771+254            & 16.9±25.6        & 583±88.4       & 15.6±23.7           & 562±86.0          \\ \midrule
e-Greedy\_linear     & 0.92±1.47        & 69.5±62.6      & 0.53±0.85          & 61.1±51.8          & 4.47±0.86        & 316±56.3       & 3.43±0.59           & 295±49.9          \\
e-Greedy\_exp        & 0.84±1.15        & 71.1±63.6      & 0.65±1.01          & 73.9±86.0          & 6.01±1.01        & 361±61.5       & 3.74±0.64           & 294±49.4          \\
Softmax\_linear      & 0.71±1.02        & 59.1±68.5      & \textbf{0.49±0.81}          & \textbf{53.8±57}            & 4.65±0.78        & 325±56.6       & 3.47±0.53           & 296±50.5          \\
Softmax\_exp         & 0.94±1.39        & 127±285        & 0.55±0.91          & 111±241            & 4.5±0.77         & 277±44.6       & 3.53±0.63           & 307±51.4          \\
TS\_accept\_better   & 0.88±1.51        & 73.2±94.4      & 0.54±0.91          & 59.4±77            & 5.45±0.92        & 341±56.9       & 4.09±0.72           & 284±47.5          \\
TS\_accept\_same     & 1.19±2.02        & 84.2±111       & 0.75±1.25          & 69.6±86.9          & 6.11±1.30        & 340±63.2       & \textbf{2.94±0.55}           & \textbf{244±39.5}          \\ \bottomrule
\end{tabular}
\caption{Primal gap (PG) percentage and primal integral (PI) at 1 hour time limit  averaged over instances from D-MIPLIB and H-MIPLIB and their standard deviations. The lower, the better (“↓”). The best performance in each dataset is bolded. Baseline, LNS and \textsc{Balans} use \textbf{SCIP} as the base solver.}
% \vspace{-0.1cm}
\label{tab:details}
\end{table*}

\section{Experiments}

What remains to be seen is the performance of \textsc{Balans} in practice. For that purpose, our main research questions are:

\begin{enumerate}

    \item[\textbf{Q1:}] What is the performance comparison between the default MIP, LNS(MIP) that commits to a single neighborhood, the state-of-the-art LNS(MIP), and our ALNS(MIP) using \textsc{Balans}? Can \textsc{Balans} achieve good performance without any offline training and explore states and neighborhoods simultaneously by adapting to the instance at hand on the fly using bandits?   
    
    % a baseline, what is the performance of LNS(MIP) applying a single neighborhood compared to default MIP? 
       
    % \item[\textbf{Q1:}] As a baseline, what is the performance of LNS(MIP) applying a single neighborhood compared to default MIP? 
    % \item[\textbf{Q2:}] How does our ALNS(MIP) compare to the baselines from LNS(MIP) and default MIP? Can we achieve good performance without any offline training and explore states and neighborhoods simultaneously on the fly via ALNS(MIP) adapting to the instance at hand? 
    \item[\textbf{Q2:}] How is arm selection among the portfolio of neighborhoods distributed in our bandit strategy? Does \textsc{Balans} depend on the single best neighborhood, or can it improve over the single best by applying weaker operators sequentially in an adaptive fashion? 
    % \item[\textbf{Q3:}] Finally, how does our online approach compare against a method that is trained offline heavily?

\end{enumerate}

% \subsection{Datasets}
% \noindent \textbf{Datasets: }
\paragraph{Datasets:}
In our experiments\footnote{https://huggingface.co/datasets/balans}, we use the recently introduced Distributional MIPLIB (D-MIPLIB)~\cite{huang2024distributional} and the commonly used MIPLIB2017-Hard (H-MIPLIB)~\cite{gleixner2021miplib}. For the former, we randomly select 10 instances from Multiple Knapsack, Set Cover, Maximum Independent Set, Minimum Vertex Cover, and Generalized Independent Set Problem, yielding 50 instances. For the latter, we consider a subset that permits a feasible solution within 20 seconds, yielding 43 instances. SCIP and Gurobi cannot solve any of these instances to optimality within 1 hour, ensuring the hardness of our benchmarks. 

% \subsection{Approaches}
% \medskip
% \noindent \textbf{Approaches:}
\paragraph{Approaches:}
For comparison, we consider the following: 

% \begin{itemize}

    % \smallskip
    % \noindent \textbf{- MIP}:
    \paragraph{- MIP:}We use \textbf{SCIP} and \textbf{Gurobi}, the state-of-the-art open-source and commercial MIP solvers with default settings running single thread~\cite{bolusani2024scip,gurobi}.

    % \smallskip
    % \noindent \textbf{- State-of-the-art LNS(MIP)}: 
    \paragraph{- State-of-the-art LNS(MIP):} We use \textbf{lb-relax} thanks to the original implementation from ~\cite{huang2023local}. This algorithm selects the neighborhood with the local branching relaxation heuristic as discussed earlier. 
    
    % \smallskip
    % \noindent \textbf{- Single Neighborhood LNS(MIP)}: 
    \paragraph{- Single Neighborhood LNS(MIP):} All eight operators  are implemented and readily available in \textsc{Balans} to serve in LNS(MIP). That said, in our experiments, we found DINS and Random Objective to perform poorly, hence we focus on the remaining six destroy operators. By varying the parameters of these operators, we obtain 16 different destroy operators from 6 unique neighborhood definitions. These are: crossover, lb\_10/25/50 (local branching), mutation\_25/50/75, proximity\_05/15/30, rens\_25/50/75, and rins\_25/50/75. For the accept criterion, we use Hill Climbing (HC) and Simulated Annealing (SA) with an initial temperature set to 20 and an end temperature set to 1 with a step size of 0.1. 
    
    % We discovered that DINS and Random Objective are not strong enough to match the hard MIP instances we consider in this paper. While all of the operators are implemented and readily available \textsc{Balans}, in the following, we omit DINS and Random Objective and consider the remaining six operators. Some of the destroy operators are parameterized, e.g., the size of the subset of variables to destroy. Varying these parameters, we obtain 16 different destroy operators from 6 unique neighborhoods, namely crossover, local\_branching\_10/25/50, mutation\_25/50/75, proximity\_05/15/30, rens\_25/50/75, rins\_25/50/75. For each one, we run Hill Climbing(HC) and Simulated Annealing(SA) seperately as accept criteria.

    % \smallskip
    % \noindent \textbf{- \textsc{Balans} ALNS(MIP)}:
    \paragraph{- \textsc{Balans} ALNS(MIP):}Given the 16 different single destroy operators used in LNS(MIP), we build \textsc{Balans} for ALNS(MIP) with a portfolio that includes all of the 16 operators. We again apply HC and SA as the acceptance criterion. For the learning policy, we use e-Greedy~\cite{auer2002finite} and Softmax~\cite{luce1977choice} with numeric rewards and Thompson Sampling (TS)~\cite{thompson1933likelihood} with binary rewards. For numeric rewards, we use a linear ([3, 2, 1, 0]) and an exponential setting ([8, 4, 2, 1]) corresponding to best, better, accept, and reject outcomes. For binary rewards, we consider [1, 1, 0, 0], where accept reward is the same as reject, and [1, 1, 1, 0], where accept is better than reject.

    % We run \textsc{Balans} with all 16 single destroy operators as our bandits portfolio. \textsc{Balans} have different configuration on MAB learning policy, scores, accept criteria. We choose Epsilon Greedy(e-Greedy), Softmax, Thompson Sampling(TS) as MAB learning policy,  [3,2,1,0](linear) and [8,4,2,1](exp) as distribution for scores, and Hill Climbing(HC) and Simulated Annealing(SA) for accept criteria. The experiments run with a grid search on all these configurations. 
% \end{itemize}

% \subsection{Evaluation Metrics}
% \medskip
% \noindent \textbf{Evaluation Metrics:} 
\paragraph{Evaluation Metrics:}
1) Primal Gap (PG)~\cite{berthold2006primal} is the normalized difference between the primal bound $v$ and a precomputed best known objective value $v^*$ and is defined as $\frac{|v-v^*|}{\max(|v^*|, \epsilon)}$ if $v$ exists and $vv^* \geq 0$. We use $\epsilon=10^{-8}$ to avoid zero division.\\ 2) Primal Integral (PI)~\cite{achterberg2012rounding} at time $q$ is the integral on $[0, q]$ of the primal gap as a function of runtime. PI captures the quality of and the speed at which solutions are found.

\begin{table*}[t]
\centering
\renewcommand{\arraystretch}{1.2}
\begin{tabular}{@{}lcccccc@{}}
\toprule
\multicolumn{1}{c}{}            & Crossover & Local Branching & Mutation & Proximity & RENS & RINS   \\ \midrule
D-MIPLIB (Softmax\_linear\_SA)  & 6.4\%     & 11\%            & 25\%     & 17.9\%    & 19.8\% & 19.9\%  \\
H-MIPLIB (TS\_accept\_same\_SA) & 9.4\%     & 0.6\%           & 21\%     & 7.3\%     & 31\% & 30.8\% \\ \bottomrule
\end{tabular}
\caption{The distribution of arm selection as a percentage for each unique destroy operator of the best \textsc{Balans} configurations.}
% \vspace{-0.3cm}
% in Distributional MIPLIB (D-MIPLIB) and MIPLIB Hard (H-MIPLIB).}
\label{tab:arms}

\end{table*}
% \subsection{Setup}
% \smallskip
% \noindent \textbf{Setup:} 
\paragraph{Setup:} We conduct experiments on AWS EC2 Trn1 with 128 vCPUs and 512GB memory. \textsc{Balans} solver integrates \textsc{ALNS} library~\cite{Wouda_Lan_ALNS_2023}, \textsc{MABwiser} library~\cite{DBLP:conf/ictai/StrongKK19,DBLP:journals/ijait/StrongKK21} and \textsc{SCIP}(v9.0.0)~\cite{bolusani2024scip} and \textsc{Gurobi} (v11.0.0)~\cite{gurobi}. 
% At every iteration of (A)LNS, we use single-threaded SCIP to solve the sub-MIP. 
To find an initial solution, we run the solver for 20 seconds. Each LNS iteration is limited to 1 minute, except for Local Branching to 2.5 minutes, which solves larger sub-problems than other operators. The time limit to solve each instance is set to 1 hour. 

% We set the time limit to 60 minutes to solve each instance and 1 minute for each repair operation in ALNS. Except for Local Branching, we set the time limit to 2.5 minutes since it solves a larger MIP than other approaches typically requires a longer time.

\subsection{Numerical Results}

% \noindent \textbf{[Q1] Default MIP vs. LNS(MIP) vs. \textsc{Balans}}\\
\paragraph{[Q1] Default MIP vs. LNS(MIP) vs. \textsc{Balans}} 
Figure~\ref{fig:lines} presents a detailed comparison of default MIP via SCIP, LNS(MIP) via single neighborhood operators including the state-of-the-art lb-relax~\cite{huang2023local}, and \textsc{Balans} solver conducting ALNS(MIP). 
%The figure shows the primal gap as a function of time, the lower the better. The top row is for the results obtained on D-MIPLIB and the bottom row is for the results obtained on H-MIPLIB. 
% The left four plots are for LNS(MIP) that commits to a single neighborhood, and the right four plots are for \textsc{Balans} conducting ALNS(MIP) with bandits. LNS(MIP) and \textsc{Balans} are run with Hill Climbing and Simulated Annealing. In each plot, the performance of the default MIP using SCIP is shown by the black line. Within each dataset (row), the black SCIP lines are identical to serve as a baseline. For LNS, each line represents a different single neighborhood. For \textsc{Balans}, each line represents a different learning policy. We only show the lines that perform within 5\% (D-MIPLIB) and 25\% (H-MIPLIB) primal gap compared to default SCIP. 

% \smallskip
% \noindent \textbf{} 
\paragraph{Performance of LNS(MIP): }On the left four plots, the immediate observation is the mixed performance of single neighborhood LNS, losing to SCIP (above the black line) in almost all cases. On D-MIPLIB, only Local Branching (lb) and Mutation get close to SCIP, but still worse. All others, including lb-relax, perform poorly above the primal gap cutoff. On H-MIPLIB, both Proximity and Local Branching are comparable to SCIP, with lb\_25 surpassing SCIP. It is important to note that these are aggregate results across problems. As expected, the state-of-the-art lb\_relax performs comparable and even better than SCIP in some domains. Domain-specific results can be found in the supplemental material. Overall, when dealing with MIP instances from various distributions and mixed problem types, single neighborhood approaches, including lb\_relax, struggle. LNS(MIP) requires carefully chosen neighborhood sizes and cannot adapt to the diversity of mixed problem types, number of variables, constraints, and problem structure across instances.

% \smallskip
% \noindent \textbf{Performance of \textsc{Balans}: } 
\paragraph{Performance of \textsc{Balans}: }Figure~\ref{fig:lines} shows that, contrary to LNS(MIP), the ALNS(MIP) approach of \textsc{Balans} significantly outperforms SCIP (below the black line) \textit{in all cases on both datasets}. In fact, it immediately starts performing better, diving into the lower optimality gap early in the search. This result is remarkable since \textsc{Balans} uses the same 16 single neighborhoods from LNS(MIP), most of which do not even reach the gap cutoff when running standalone. An important insight is that the ranking of single LNS(MIP) operators varies considerably between datasets. For instance, Mutation performs the best on D-MIPLIB yet is absent in H-MIPLIB, whereas Proximity exhibits the opposite behavior. This is exactly the motivation behind our diverse portfolio of destroy operators. Notice also how \textit{all bandit learning policies} are consistently better than SCIP. Our online bandits are able to select the appropriate operator adaptively throughout the search for the instance at hand. Moreover, unlike single operators that require carefully tuned destroy parameters, in the absence of which they perform poorly, \textsc{Balans} simply introduces parameter variations as additional arms driven by the learning policy.

Table~\ref{tab:details} zooms further into the primal gap and primal integral achieved by the top performers. As in Figure~\ref{fig:lines}, it is evident that \textit{any configuration} of \textsc{Balans} 
significantly outperforms SCIP, the state-of-the-art lb-relax, and any LNS. Overall, we reduce the primal gap of SCIP by \textbf{75+\% } and the primal integral of SCIP by \textbf{50+\% }across datasets. The best \textsc{Balans} configuration is Softmax\_linear\_SA for D-MIPLIB and TS\_accept\_same\_SA for H-MIPLIB. Regardless, any \textsc{Balans} configuration is better than the alternatives, revealing its robust out-of-the-box performance.

% Figure 2 (and Table 2) provides a comparison of our \textsc{Balans} method across different configurations. 

% Within the columns of Table 2, Simulated Annealing (SA) consistently outperforms Hill Climbing (HC) as an acceptance criterion. This indicates that SA contributes more effectively to the \textsc{Balans} solver's performance than it does to single heuristics, thereby yielding even better results. 

\begin{figure}[t]
   \centering
   \includegraphics[width=\linewidth]{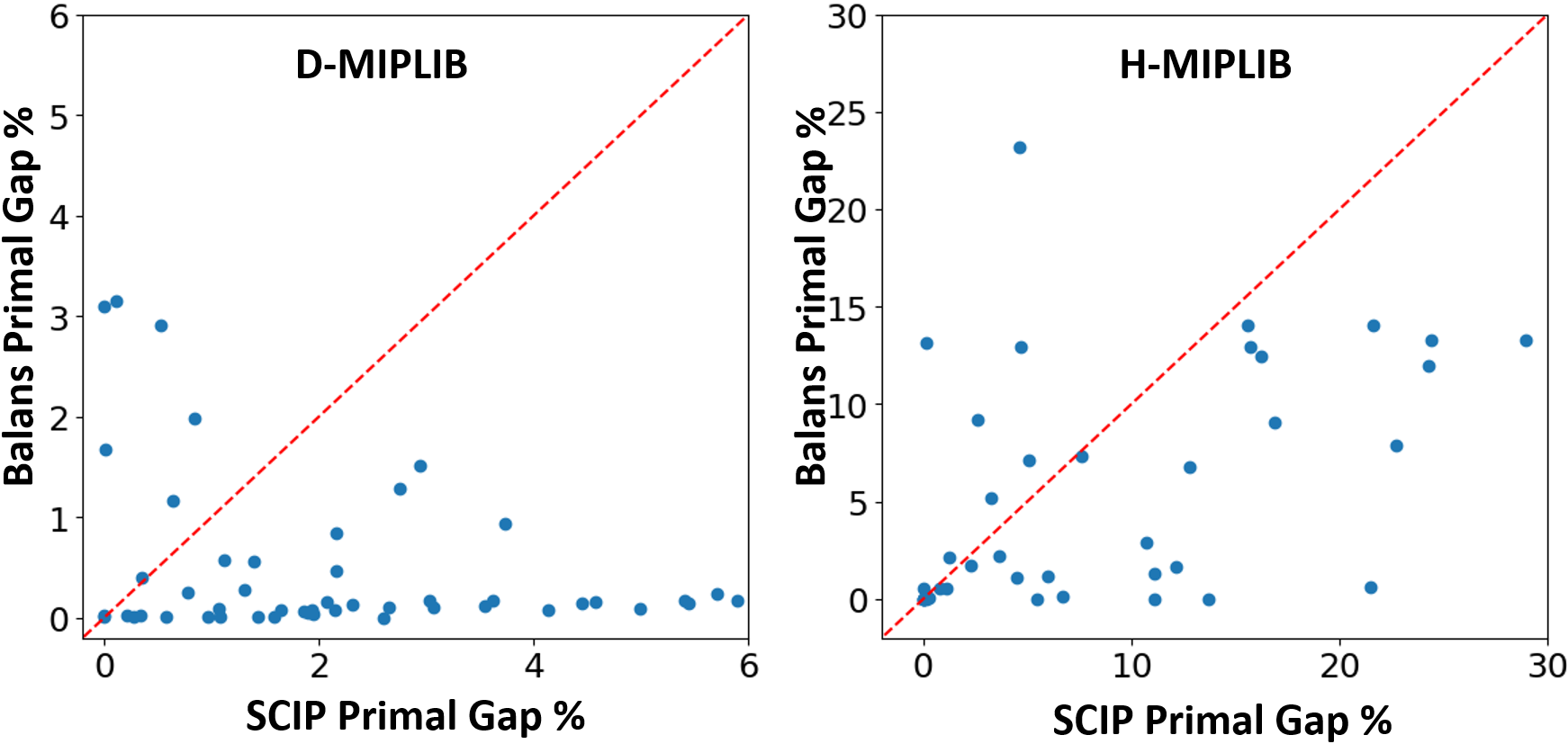}
% \vspace{-0.5cm}
\caption{Primal gap of SCIP (x-axis) vs. \textsc{Balans} (y-axis).
% for D-MIPLIB (left) and H-MIPLIB (right).
}
\label{fig:scatter}
% \vspace{-0.3cm}
\end{figure}

Figure~\ref{fig:scatter} plots the primal gap performance of the best \textsc{Balans} configuration within each dataset compared to SCIP on each instance. 
Above the diagonal is where SCIP wins, and below the diagonal is where \textsc{Balans} wins. As clearly shown in Figure~\ref{fig:scatter}, \textsc{Balans} achieves superior performance on a greater number of instances, particularly on those where SCIP struggles. This highlights the effectiveness of our approach in challenging instances.

% \smallskip
% \noindent \textbf{Performance of Acceptance Criteria: } 
\paragraph{Performance of Acceptance Criteria: }For single operators in LNS(MIP), different acceptance criteria do not substantially change the performance. They can be considered too rigid to benefit from state exploration. In contrast, SA consistently outperforms HC for \textsc{Balans}. The best \textsc{Balans} performances are achieved with SA that allows navigating the search space even with worsening moves. 

% Interestingly, there is no single configuration of \textsc{Balans} that consistently outperforms the others across different datasets, underscoring the adaptability and necessity of our \textsc{Balans} approach. Nonetheless, even the least effective \textsc{Balans} configuration still surpasses SCIP and single heuristics, demonstrating that any \textsc{Balans} configuration can be reliably applied to problem instances without risking poor performance.

% Note that we did not tune for the \textsc{Balans} portfolio. A careful portfolio construction and algorithm selection are interesting future work. Here, we show that out-of-the-box performance of \textsc{Balans} outperforms default MIP, any single LNS(MIP) and the best LNS(MIP) from previous literature. 

% \noindent \textbf{[Q2] Distribution of Arm Selection} 
\paragraph{[Q2] Distribution of Arm Selection} What is behind the operations of \textsc{Balans}, and specifically, what neighborhoods does it depend on for its superior performance? 

 Table~\ref{tab:arms} shows the distribution of arm selection as percentages. Interestingly, the single best operator, local branching, is \textit{not} a popular arm at all. Several worse operators are preferred within our bandit portfolio. In H-MIPLIB, \textsc{Balans} utilizes local branching \textit{only 0.6\% }of the time, yet still, it outperforms local branching by a large margin. Likewise, in D-MIPLIB, local branching is second to last in popularity. 
 
 The RENS and RINS operators, which are not even visible in Figure~\ref{fig:lines} due to their poor performance above the primal gap cutoff, account for ${\sim}40$\% and ${\sim}60$\% of the distributions in Table~\ref{tab:arms} on these datasets. This shows that \textsc{Balans} outperforms MIP and any single best LNS(MIP) out-of-the-box with almost no tuning by \textit{using weaker operators sequentially} and carefully balancing the exploitation-exploration trade-off via online learning.

% \vspace{-0.1cm}
\subsection{Ablation Studies}
To demonstrate that \textsc{Balans} is solver-agnostic, we conducted the same experiments using Gurobi~\cite{gurobi}, a state-of-the-art commercial solver. 

\begin{figure}[t]
   \centering
   \includegraphics[width=\linewidth]{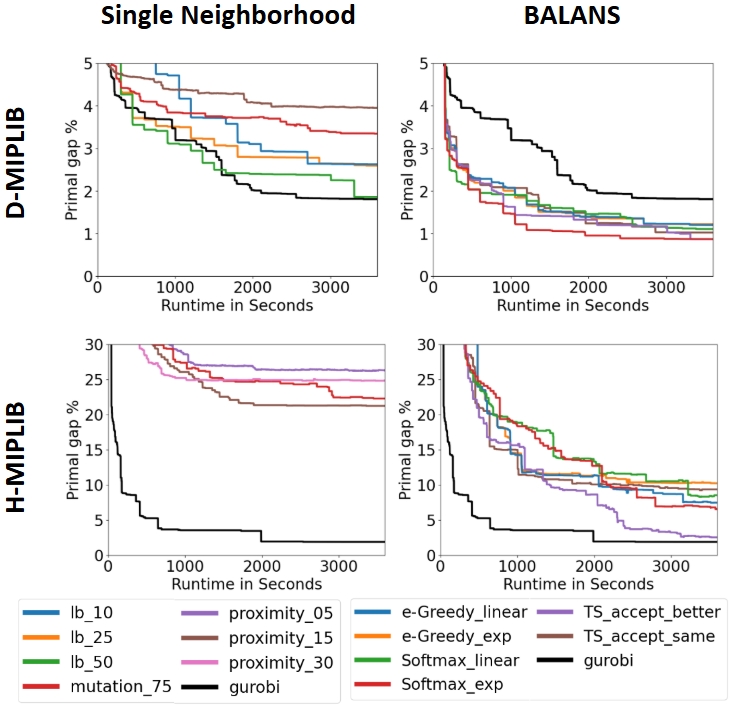}
% \vspace{-0.5cm}
\caption{Ablation Study: The same experiments from Figure \ref{fig:lines} using Gurobi as the MIP solver with Simulated Annealing shown for the acceptance criterion.
% for D-MIPLIB (left) and H-MIPLIB (right).
}
\label{fig:gurobi}
% \vspace{-0.3cm}
\end{figure}

Figure~\ref{fig:gurobi} compares the performance of default Gurobi, the Single Neighborhood LNS approach, and \textsc{Balans}. The results show that \textsc{Balans} outperforms both default Gurobi and LNS(Gurobi) with any single neighborhood approach significantly on D-MIPLIB. On H-MIPLIB, where Gurobi perform much better than any LNS methods, our \textsc{Balans} can significantly reduce the gap from single LNS to Gurobi. This confirms that our findings and conclusions derived from experiments with the SCIP solver carry over when switched to another solver. By conducting experiments with state-of-the-art open-source and commercial solvers, we demonstrate that \textsc{Balans} can be seamlessly applied as a meta-solver on top of a MIP solver, significantly improving their performance over single neighborhood LNS. 

% Notice also that we use the same 16 neighborhoods in both experiments, i.e., the portfolio of \textsc{Balans} is not explicitly tuned per solver. By conducting experiments with state-of-the-art open-source and commercial solvers, we demonstrate that \textsc{Balans} can be seamlessly applied as a meta-solver on top of a MIP solver, significantly improving their performance. The complete details of experiments using Gurobi are in Appendix C.

Next, it is important to note that the default configuration of SCIP applies ALNS as an internal heuristic \textit{within} the Branch-and-Bound search, as discussed in~\cite{hendel2022adaptive}. That means when we use the default SCIP, \textsc{Balans} essentially conducts ALNS(MIP(ALNS)), which outperforms the default SCIP conducting MIP(ALNS). 
% Our experimental results show that \textsc{Balans} significantly outperforms default SCIP, indicating that ALNS(MIP) is superior to MIP(ALNS). 
This raises the question: how much of the good performance of \textsc{Balans} on these instances is due to the internal ALNS of SCIP? To explore this, we conduct another ablation study with SCIP's internal ALNS enabled and disabled, comparing MIP vs. MIP(ALNS). Our results show that the internal ALNS of SCIP conducted within the BnB search led to performance improvement in only 8 out of 94 instances, with negligible improvements of less than 0.2\% in the primal gap. Contrarily, our novel bandit-based ALNS approach, conducted on top of the BnB search, improves \textbf{79 out of 94} instances (Figure~\ref{fig:scatter}) and reduces the primal gap by \textbf{75+\%} (Table\ref{tab:details}).

% TODO compare ALNS(MIP) vs. ALNS(MIP(ALNS)

% \vspace{-0.2cm}
\section{Related Work}
% A growing body of research is dedicated to integrating machine learning and mixed-integer optimization. 
% These include general algorithm configuration procedures~\cite{DBLP:conf/cp/KadiogluMSSS11}, variable selection~\cite{khalil2016learning}, cut selection~\cite{paulus2022learning}, node selection\cite{he2014learning,labassi2022learning}, and theoretical results for tree-search configuration~\cite{DBLP:conf/nips/BalcanPSV21,DBLP:conf/cp/BalcanPSV22}.
% ~\cite{khalil2016learning,BENGIO2021405,DBLP:conf/aaai/KadiogluMS12,dash,lodi2017learning,DBLP:conf/cp/KadiogluMSSS11}

Primal heuristics are crucial in rapidly discovering feasible solutions to improve the primal bound. As such, there is a rich literature on learning-based methods for improving heuristics. Examples in this line of research include IL-LNS~\cite{sonnerat2021learning}, which learns to select variables by imitating LB, RL-LNS~\cite{wu2021learning}, which uses a similar framework but trained with reinforcement learning, and CL-LNS~\cite{huang2023searching} uses contrastive learning to learn to predict variables. Our work complements these approaches as an \textit{online method} that does not require any offline training to generate good quality solutions on challenging optimization instances, as shown in our experiments. 

% Examples in this line of research include~\cite{ding2020accelerating}, which uses a local branching constraint to choose a subset of the binary variables to predict partial solution, ~\cite{nair2020solving, khalil2022mip}, which fix  variables to predicted values as partial assignment to guide a MIP solver, and~\cite{huangcontrastive}, which uses contrastive learning for variable assignments and then solves the resulting sub-MIP to find high-quality solutions. 

% For learning to predict and search for heuristics, ~\cite{ding2020accelerating} use a local branching constraint choose a subset of the binary variables to predict partial solution. 
% ~\cite{nair2020solving, khalil2022mip} propose to fix the variables to their predicted value and hand over this partial assignment to an MIP solver that optimizes over the remaining variables. ~\cite{huangcontrastive} use contrastive learning for predicting variable assignments and then solve the resulting reduced MIP to find high-quality solutions.

There is also a body of work that focuses on primal heuristic operations of BnB \textit{within-the-solver}. These include offline learning methods, such as~\cite{khalil2017learning}, which builds a mapping between BnB nodes and a binary decision for running a given primal heuristic, and~\cite{chmiela2021learning}, which constructs a schedule that assigns priority and computational budget to each heuristic. 

When considering online methods~\cite{attieh2019athanor,hendel2022adaptive,chmiela2023online} are the closest to our work. As in our paper, both are examples of online learning, focus on LNS neighborhoods, use SCIP, and are based on multi-armed bandits. The crucial difference is that \textsc{Balans} is a \textit{meta-solver} that operates on top of a MIP solver in a solver-agnostic manner, whereas the previous work only works as a heuristic that operates within the BnB search of SCIP, where LNS is invoked at internal search nodes. As a result, these works remain specific extensions of SCIP and are white-box, i.e., require access to solver internals. Contrarily, \textsc{Balans} is an integration technology that offers a modular architecture to leverage best-in-class open-source software dedicated to their specific domain. We leverage  \textsc{MABwiser}~\cite{DBLP:conf/ictai/StrongKK19,DBLP:journals/ijait/StrongKK21} for bandits, \textsc{ALNS} library~\cite{Wouda_Lan_ALNS_2023} for adaptive large-neighborhood search, and \textsc{SCIP}~\cite{bolusani2024scip} and \textsc{Gurobi}~\cite{gurobi} for MIP solving. Future advances in these distinct fields, realized independently within each software, propagate to our meta-solver with compounding effects. This unique integration makes \textsc{Balans} highly configurable. Beyond this paper, many more configurations are available through  parameterization of ALNS and MABWiser. 

% Moreover, our design enables the introduction of new destroy and repair operators with ease. %Each operator is defined on its own independently from the core engine and other operators. 
% Extensions become available as new parameters that plug and play with existing options. In Appendix A~\cite{cai2024balans}, we present a quick start usage example of \textsc{Balans} showcasing its ease-of-use and high-level API design. 

% A detailed performance comparison between offline trained methods and our online approach requires access to trained artifacts and is beyond the scope of this paper. We take results of offline learning approaches from~\cite{huang2023searching} as-is and compare their results \textit{directionally} with \textsc{Balans} on the same D-MIPLIB distributions. We find \textsc{Balans} perform comparably with offline learning approaches, relying only on online learning. The full table of results and analysis is in the appendix.
% The best performing offline method, CL-LNS, from~\cite{huang2023searching} trains a supervised approach based on contrastive learning to predict the destroy neighborhood. Comparing their results \textit{directionally} on the same D-MIPLIB instances based on Table 8 from that paper, \textsc{Balans} performs competitively, relying only on online learning.

Finally, let us acknowledge other successful MAB applications. Among those,~\cite{phan2024adaptive} uses MAB with ALNS for multi-agent pathfinding,~\cite{almeida2020hyper} uses MAB with hyper-heuristics for multi-objective flow shop problems, and ~\cite{zheng2022bandmaxsat} uses MAB for MaxSAT. Beyond combinatorial problems, MAB is heavily used in recommender systems~\cite{DBLP:conf/aaai/KadiogluK24} and game-playing agents~\cite{schaul2019adapting}. 
% Our work is another addition to this growing literature.

% \vspace{-0.1cm}
\section{Conclusions}
In this paper, we proposed \textsc{Balans}, a multi-armed \textbf{B}andits-based \textbf{A}daptive \textbf{La}rge \textbf{N}eighborhood \textbf{S}earch for mixed-integer programming. Our experiments have shown that all configurations of our online approach, without any offline training and almost zero tuning, significantly improve over 1) the default BnB solver,  and 2) any single large neighborhood search, including the state-of-the-art, on hard optimization problems. We release \textsc{Balans} solver as open-source software with its high-level interface, modular and extendable architecture, and configurable design. 

For future work, we would like to improve the performance of \textsc{Balans} further by careful algorithm configuration and portfolio construction~\cite{DBLP:conf/ecai/KadiogluMST10,DBLP:conf/cp/KadiogluMSSS11}, and explore hybrid ALNS(MIP), where existing offline training methods are introduced as additional arms in our portfolio. 
\section*{Acknowledgments}
This paper reports on research conducted while Junyang Cai was interned at AI Center of Excellence, Fidelity. The research at the University of Southern California was supported by NSF under grant number 2112533:NSF Artificial Intelligence Research Institute for Advances in Optimization (AI4OPT).

\bibliographystyle{named}
\bibliography{ijcai25}
\newpage

\appendix
\onecolumn

\section*{Appendix}
In this Appendix, we present a quick start example of \textsc{Balans} showcasing its ease-of-use and high-level API design, a detailed explanation of our destroy operators and how they differ from their original reference, the detailed results on each problem distribution, and finally, detailed results and analysis about offline and online training comparison.

\section{\textsc{Balans} Quick Start Example}
% \medskip

\begin{minted}[fontsize=\footnotesize]{python}
# Adaptive large neigborhood via ALNS library
from alns.select import MABSelector
from alns.accept import HillClimbing
from alns.stop import MaxIterations

# Contextual multi-armed bandits via MABWiser library
from mabwiser.mab import LearningPolicy

# Meta-solver for MIPs via SCIP library
from balans.solver import Balans, DestroyOperators, RepairOperators

# Balans: Online Meta-Solver for MIPs
balans = Balans(destroy_ops=[DestroyOperators.Crossover,
                             DestroyOperators.Dins, 
                             DestroyOperators.Mutation, 
                             DestroyOperators.Local_Branching,
                             DestroyOperators.Proximity,
                             DestroyOperators.Rens, 
                             DestroyOperators.Rins,
                             DestroyOperators.Random_Objective],
                repair_ops=[RepairOperators.Repair],
                selector=MABSelector(scores=[1, 1, 0, 0], num_destroy=8, num_repair=1,   
                                     learning_policy=LearningPolicy.ThompsonSampling()),
                accept=HillClimbing(),
                stop=MaxIterations(100))

# Run on MIP instances
result = balans.solve("MIP_instance.mps")

# Result for the best solution found
print("Best solution:", result.best_state.solution())
print("Best solution objective:", result.best_state.objective())
\end{minted}

\medskip

As shown in our Quick Start Example, \textsc{Balans}\footnote{https://github.com/anonymous/balans/} solver is characterized by a set of \textit{destroy operators} and a set of \textit{repair operators}. The pairwise combination of these two is treated as an \textit{arm} in the multi-armed bandits, with the option to declare valid arm pairs if needed. A \textit{selector} controls the selection among the arms. The selector uses a given learning policy, which is rewarded for finding the best, better, accepted, and rejected state according to the rewards given in the \textit{scores} list. The ALNS iterates following the \textit{acceptance criterion} and terminates with the \textit{stopping condition}.

\medskip
We contribute \textsc{Balans} to the open-source as a meta-solver for solving MIPs. It serves as an integration technology that combines \textsc{ALNS} library for adaptive large-neighborhood search, \textsc{MABwiser} library for multi-armed bandits, and \textsc{SCIP} library for solving MIPs. Our integration yields an elegant interface with several configurable options, from reward mechanisms and learning policies to acceptance criteria. The framework is available in the Python Package Index (PyPI) and can be installed with a one-liner\footnote{pip install balans}.

\medskip

\textsc{Balans} provides eight built-in destroy operators and one repair operator. It takes advantage of \textit{any learning policy} available in  \textsc{MABWiser} library and a\textit{ny acceptance and stopping criteria} available in \textsc{ALNS} library. Our modular software leverages best-in-class open-source software dedicated to their specific algorithms for bandits, adaptive large-neighborhood search, and MIP solving. Further advances in these distinct domains realized independently in each software propagate to enhancing and compounding our \textsc{Balans} meta-solver.

\newpage
\section{The Details of LNS and Destroy Operators}
In this section, we provide more details on the LNS procedure and the details of our destroy operators as implemented in \textsc{Balans}. The general LNS procedure is as follows:
\begin{enumerate}
    \item Given a MIP, (A)LNS starts with a feasible solution. That is, we have a ``complete assignment" to all variables and the corresponding value of the objective function. This is called a ``state".
    \item Then, when a Destroy operator is applied to a state, it takes this ``complete assignment" and returns a sub-MIP problem. Essentially, the destroy operation transitions from a complete solution back to an unsolved MIP problem. The main motivation is that this sub-problem is easier than solving the original MIP.
    \item Finally, a Repair operator takes an unsolved sub-MIP and solves/repairs it. This repair leads to another ``complete assignment" of variables. This forms the next state. An Acceptance Criterion decides whether we move from the completed assignment (solution) of the current state to the complete assignment (solution) of the next state. 
    \item The interplay between the Destroy/Repair cycle is repeated until a Stop Criterion is encountered. 
\end{enumerate}

\begin{table*}[t]
\centering

% \begin{tabular}{c|c|c}
\begin{tabular}{@{}ccc@{}}
\toprule
                 & Destroy Neighborhood Definition & Added Constraint(s) \\ \midrule
Crossover        & $\textit{Destroy} = \{x_{t-1}^{k} == x_{rnd}^{k}, \forall k \in D\}$   &  $\{x_{t}^{k} = x_{t-1}^{k} \mid k \notin \textit{Destroy}, \forall k \in V\}$  \\ \midrule
DINS (IP)             &  $\textit{Destroy} =\{|x_{t-1}^{k} - x_{lp}^{k}| \geq 0.5, \forall k \in I\}$  &     $ \begin{cases}
        |x_{t}^{k} - x_{lp}^k| \leq |x_{t-1}^{k} - x_{lp}^k| & \mid k \in \textit{Destroy} \\
        x_{t}^{k} = x_{t-1}^{k} & \mid k \notin \textit{Destroy}
    \end{cases}  , \forall k\in V\bigg\}$   \\  \midrule
Local Branching (BP)  &   $\textit{Destroy} = \Delta B$        &      $\sum\limits_{k\in \textit{B}} |x_{t}^k - x_{t-1}^k| \leq |\textit{Destroy}|$              \\  \midrule
Mutation         &  $\textit{Destroy} =\Delta D $       &   $\{x_{t}^{k} = x_{t-1}^{k} \mid k \notin \textit{Destroy}, \forall k \in V\}$                                     \\ \midrule
Proximity Search (BP) &    $f'(x_t) = \sum\limits_{k\in \textit{B}} |x_{t}^k - x_{t-1}^k| $       &      $f(x_{t}) \leq \Delta f(x_{t-1}) $                                \\ \midrule
Random Objective &   $f'(x_t) = \sum\limits_{k\in V} {\Delta x_{t}^k} $       &                   -                    \\ \midrule
RENS             &  $\textit{Destroy} =\Delta\{x_{lp}^{k} \notin \mathbb{Z}, \forall k \in D\}$          &    $\begin{cases}
        x_{t} \geq \lfloor x_{lp}^k\rfloor, x_{t} \leq \lceil x_{lp}^k\rceil  & \mid k \in \textit{Destroy} \\
        x_{t}^{k} = x_{t-1}^{k} & \mid k \notin \textit{Destroy}
    \end{cases}  , \forall k\in V\bigg\}$     \\ \midrule
RINS             &  $\textit{Destroy} =\Delta \{x_{t-1}^{k} == x_{lp}^{k}, \forall k \in D\}$   &  $\{x_{t}^{k} = x_{t-1}^{k} \mid k \notin \textit{Destroy}, \forall k \in V\}$  \\ \bottomrule
\end{tabular}
% \vspace{-0.2cm}
\caption{Eight destroy operators in \textsc{Balans} with their destroy set of variables, modified objective, and constraints. We have a single repair operator that re-optimizes the MIP resulting from applying one of these destroy neighborhoods to the current state. The parentheses next to the operator name indicate the MIP problem type for which the operator is suitable for. Let $x_{t-1}^{n}$ be the solution from previous state, $x_{t}^{n}$ be the solution for current state, $x_{lp}^{n}$ be the LP relaxation solution, and $x_{rnd}^{n}$ be a random feasible solution. Let $V$ denote list of variable indices, $B$ denote list of binary indices, $I$ denote list of integer indices, and $D$ denote list of discrete indices. Let $f(x)$ to denote the objective function. The $\Delta$ is a hyper-parameter that defines either the size of the destroy set, the coefficient of variables, or the objective function.}
% \vspace{-0.4cm}
\label{tab:destroy}
\end{table*}

The destroy operators implemented in \textsc{Balans} with our modifications are as follows and given in Table \ref{tab:destroy}.

\begin{itemize}
    \item \textbf{Crossover}: Crossover generates a random feasible solution using Random Objective and solution from the previous state to compare the value of discrete variables between them. If they have the same value, fix the variable to that value. This differs from the original, which takes two random feasible solutions and compares them against each other. Using the incumbent solution allows this approach to have connections to the current state information.
    \item \textbf{DINS}: DINS takes the LP relaxed solution of the original MIP and solution from the previous state to record the difference of value between each discrete variable. If the difference is less than 0.5 for each discrete variable, we fix the variables to the previous state value. Otherwise, we bound the variables around the initial LP relaxed solution. In this approach, we remove the binary local branching part because we want each operator to be unique enough.
    \item \textbf{Local Branching}: Local Branching tracks the value of binary variables in the previous state and only allows a fraction of the binary variables to flip by adding one single constraint to the original MIP that allows $\leq k$ to change value and solving this modified MIP maximizing the original objective. We add a delta variable to control how many percent of variables we allow to flip at most. We select a random flip size between 10\% to the max percentage in every iteration. Instead of having a neighborhood with a fixed number of destroyed variables, we allow randomness in the neighborhood in every iteration.
    \item \textbf{Mutation}: Mutation fixes a subset of the discrete variables to the value from the previous state. We add a delta variable to control what percent of the discrete variables is not in the subset to be fixed.
    \item \textbf{Proximity Search}: Proximity Search finds a feasible solution with a better objective that is as close as the previous state's solution. It adds a constraint to make the objective function smaller than the previous one, replaces the objective function with the distance function between the current and previous state solutions, and minimizes the new objective. We add a delta variable to control what percent of improvement we want from the previous state. We also added a Slack variable in the constraint and objective to prevent the modified problem from becoming infeasible and make it fit in the ALNS context.
    \item \textbf{Random Objective}: Random Objective is a new heuristic we create to explore the problem space's feasible region randomly. We replace the objective function by generating a random coefficient from -1 to 1 for every variable in the problem. This acts as a single heuristic and works as a helper function for random feasible solutions in crossover and initial starting points for ALNS.
    \item \textbf{RENS}: RENS fix variables with an integer LP relaxation value to the previous state value. For variables with a fractional LP relaxation value, it restricts them from rounding up and down integer versions of that fractional value. To introduce randomness in this heuristic, we introduce a delta variable to control the percentage of floating variables to be rounded.
    \item \textbf{RINS}: RINS takes the LP relaxation solution of the original MIP and the solution from the previous state to compare the value of discrete variables between them. We choose a subset of variables with different values and fix all other variables not in the set. We introduce a delta variable to control how many percentages of variables have different values in the subset.
\end{itemize}

\begin{figure}[t]
    \centering
    \includegraphics[width=0.9\linewidth]{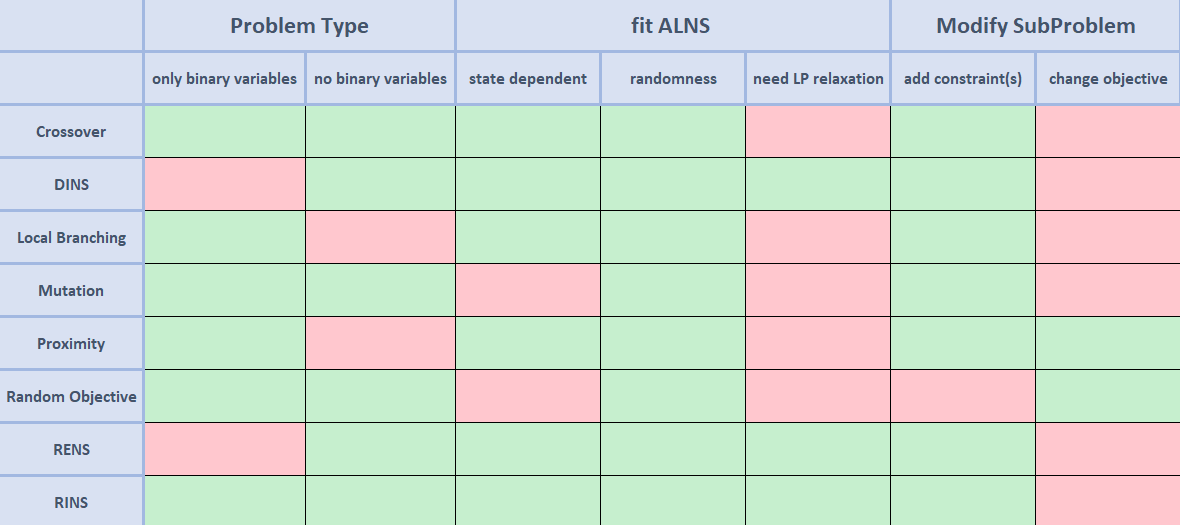}
    \caption{Different characteristics for each operator, where green means satisfy and red means not satisfy.}
    \label{fig:characteristics}
\end{figure}

Combining these eight destroy heuristics creates a diverse portfolio, including state-dependent operators and not state-independent, random exploration operators and exploration close to the previous state neighborhood, binary problem operators, and general discrete operators. Figure~\ref{fig:characteristics} presents the characteristics of each destroy operator.

\newpage

\section{Detailed Performance in Each Problem Domain}
The following Figures 2-6 show the performance of single heuristics and \textsc{Balans} on different distributions in D-MIPLIB: Multiple Knapsack (MK), Set Cover (SC), Generalized Independent Set (GISP), Minimum Vertex Cover (MVC), Maximum Independent Set (MIS). The set of experiments is done in SCIP as base solver for default solver, single heuristic and \textsc{Balans}.

The figures on the Left show the performance of single heuristics, and the figures on the Right show the performance of \textsc{Balans}. All the approaches use Simulated Annealing as accept criteria. As can be seen from these results, the performance of single heuristics varies considerably across different distributions. Meanwhile, our \textsc{Balans} solver, composed of these same operators in a bandit portfolio, performs well on \textit{every distribution}.

\begin{figure*}[ht!]
    \begin{subfigure}{.49\textwidth}
    \centering
    \includegraphics[width=.95\linewidth]{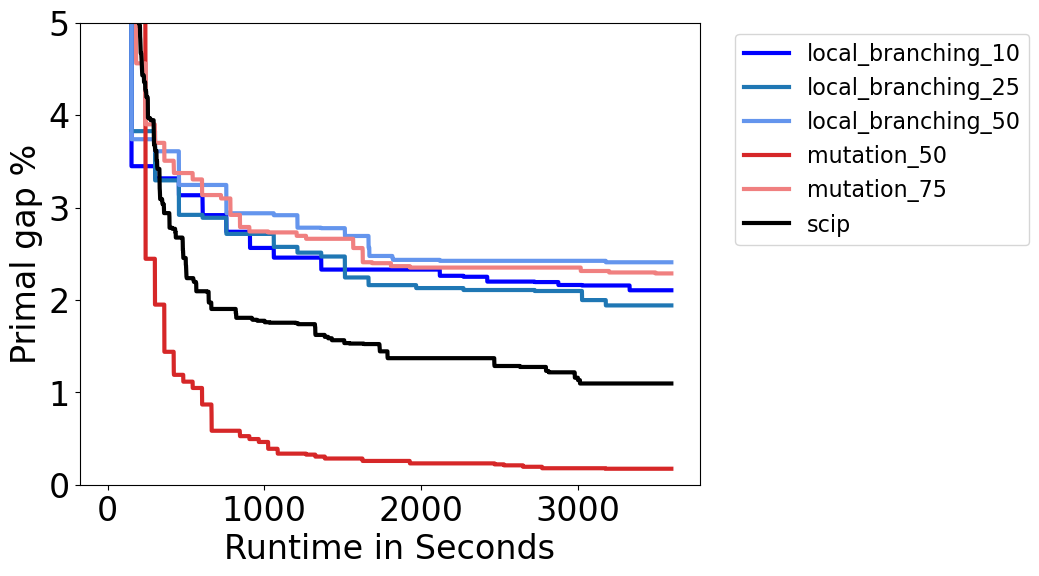}  
\end{subfigure}
\begin{subfigure}{.49\textwidth}
    \centering
    \includegraphics[width=.95\linewidth]{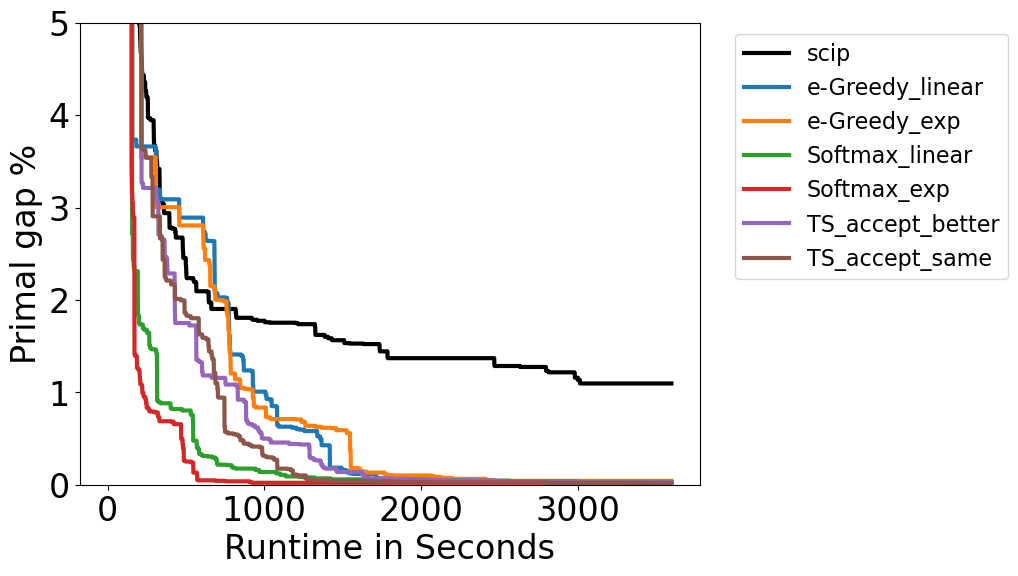}  
\end{subfigure}
\caption{The primal gap (the lower, the better) as a function of time, averaged over instances from MK distribution}
\end{figure*}

\begin{figure*}[ht!]
    \begin{subfigure}{.49\textwidth}
    \centering
    \includegraphics[width=.95\linewidth]{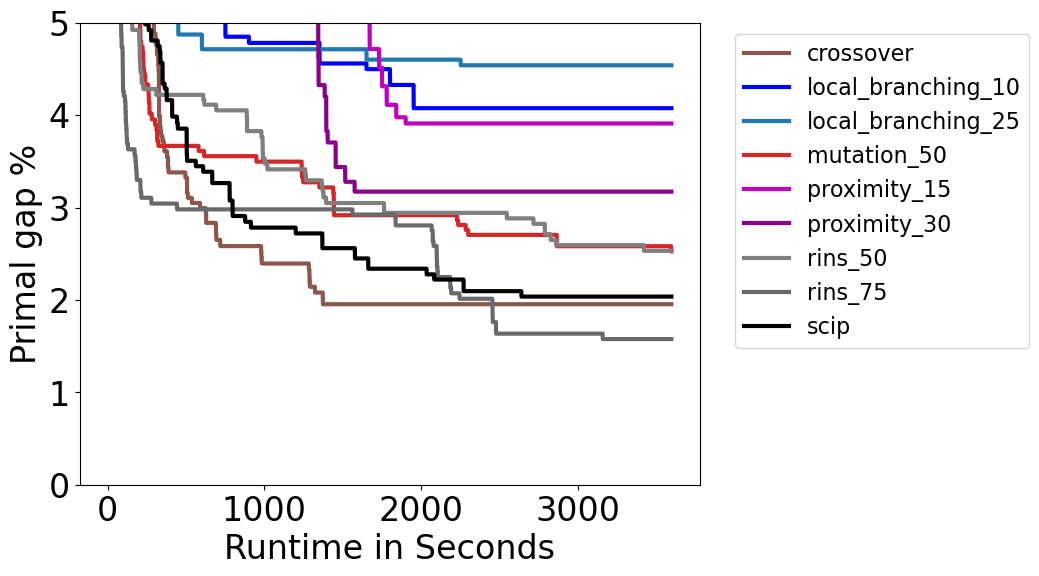}  
\end{subfigure}
\begin{subfigure}{.49\textwidth}
    \centering
    \includegraphics[width=.95\linewidth]{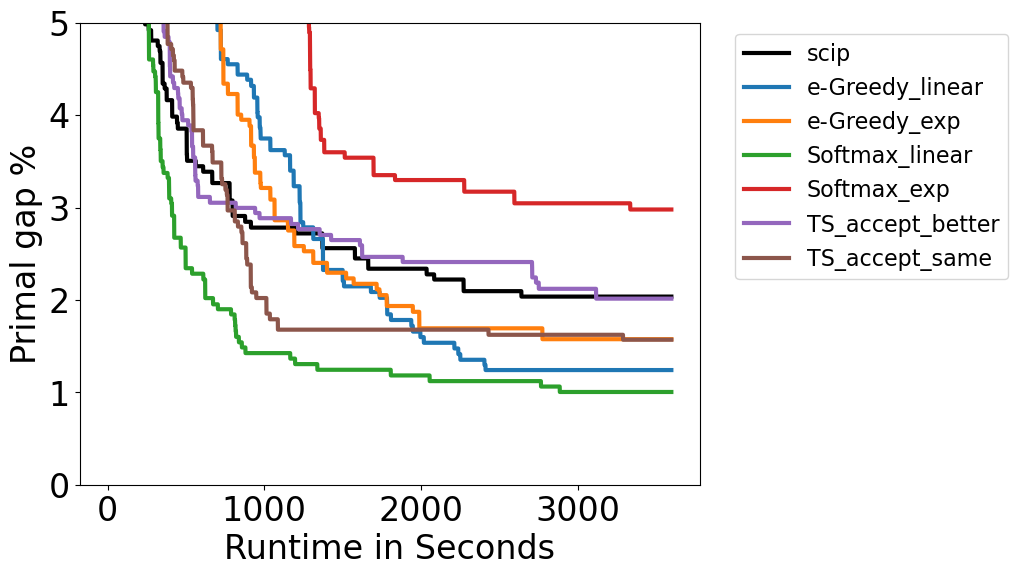}  
\end{subfigure}
\caption{The primal gap (the lower, the better) as a function of time, averaged over instances from SC distribution}
\end{figure*}

\begin{figure*}[ht!]
    \begin{subfigure}{.49\textwidth}
    \centering
    \includegraphics[width=.95\linewidth]{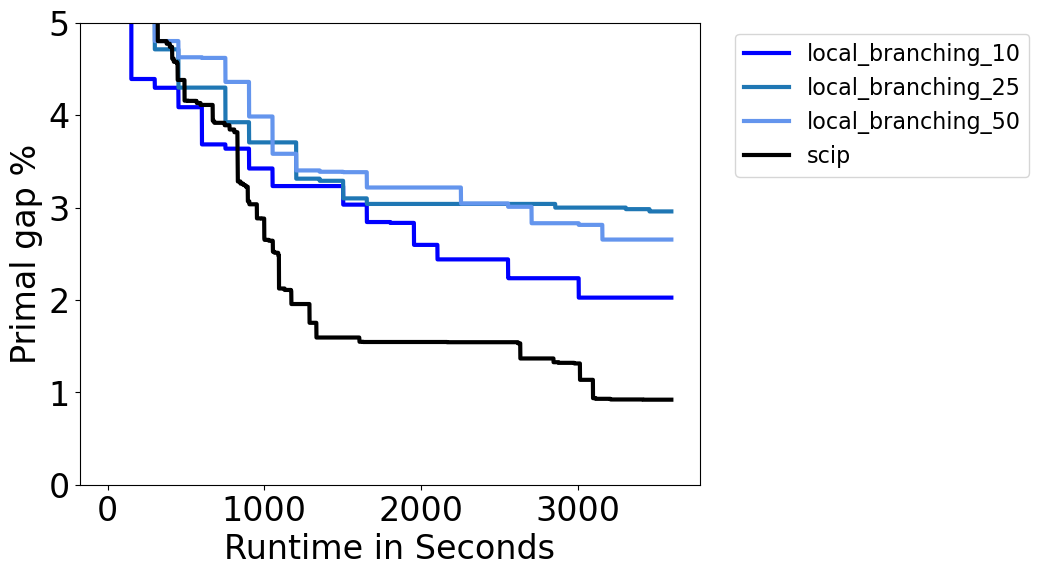}  
\end{subfigure}
\begin{subfigure}{.49\textwidth}
    \centering
    \includegraphics[width=.95\linewidth]{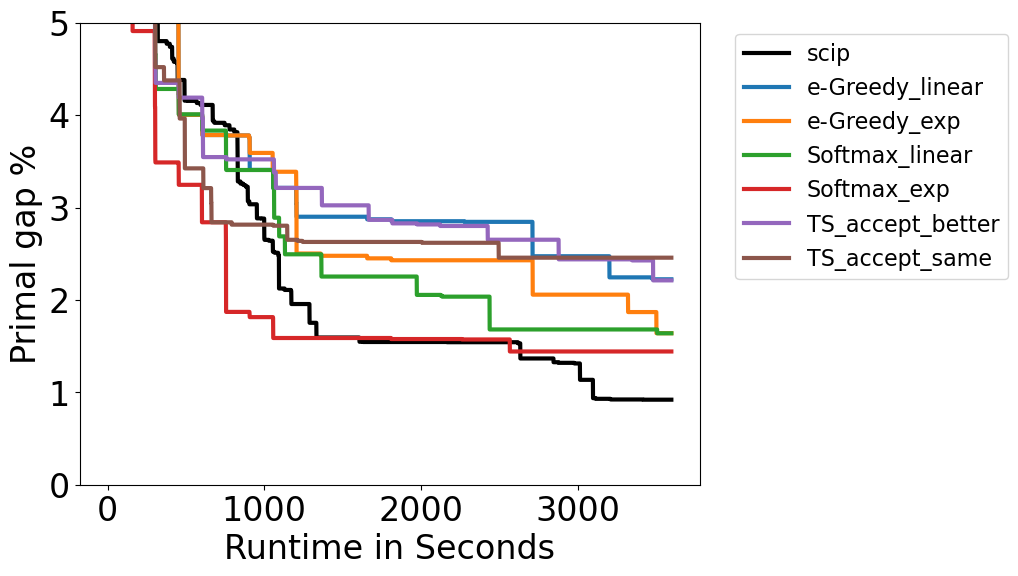}  
\end{subfigure}
\caption{The primal gap (the lower, the better) as a function of time, averaged over instances from GISP distribution}
\end{figure*}

\begin{figure*}[ht!]
    \begin{subfigure}{.49\textwidth}
    \centering
    \includegraphics[width=.95\linewidth]{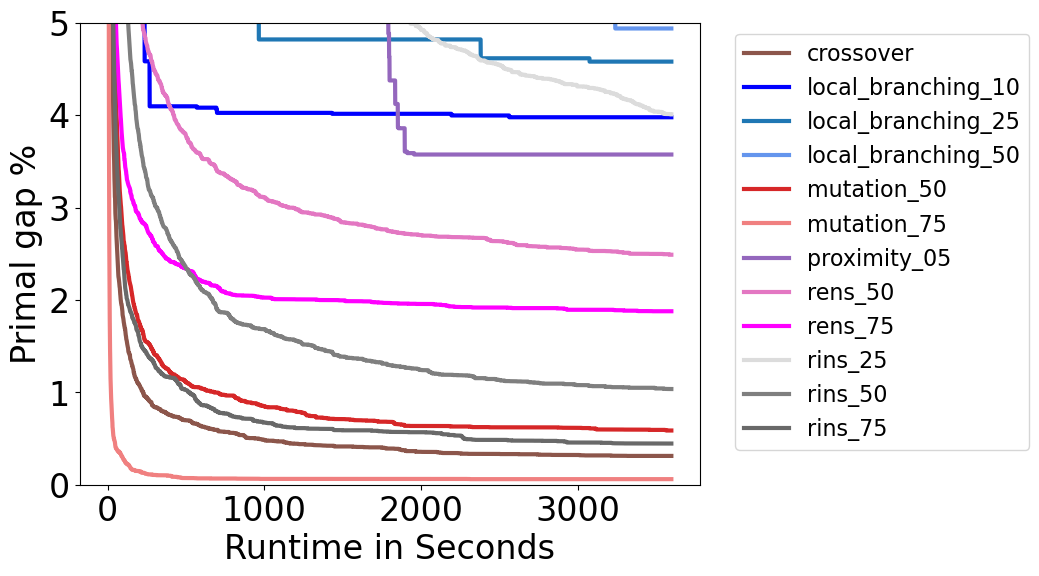}  
\end{subfigure}
\begin{subfigure}{.49\textwidth}
    \centering
    \includegraphics[width=.95\linewidth]{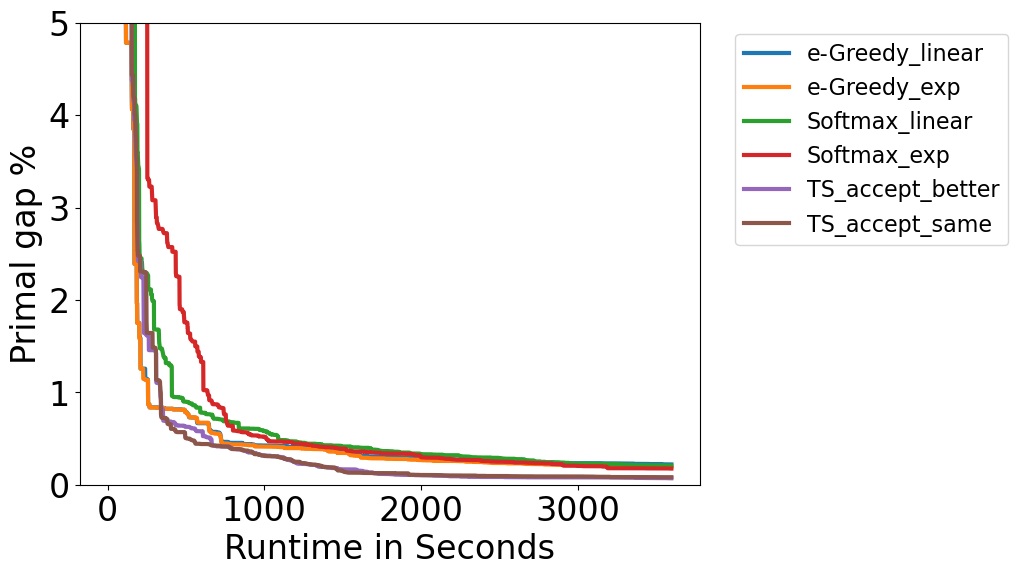}  
\end{subfigure}
\caption{The primal gap (the lower, the better) as a function of time, averaged over instances from MVC distribution}
\end{figure*}

\begin{figure*}[ht!]
    \begin{subfigure}{.49\textwidth}
    \centering
    \includegraphics[width=.95\linewidth]{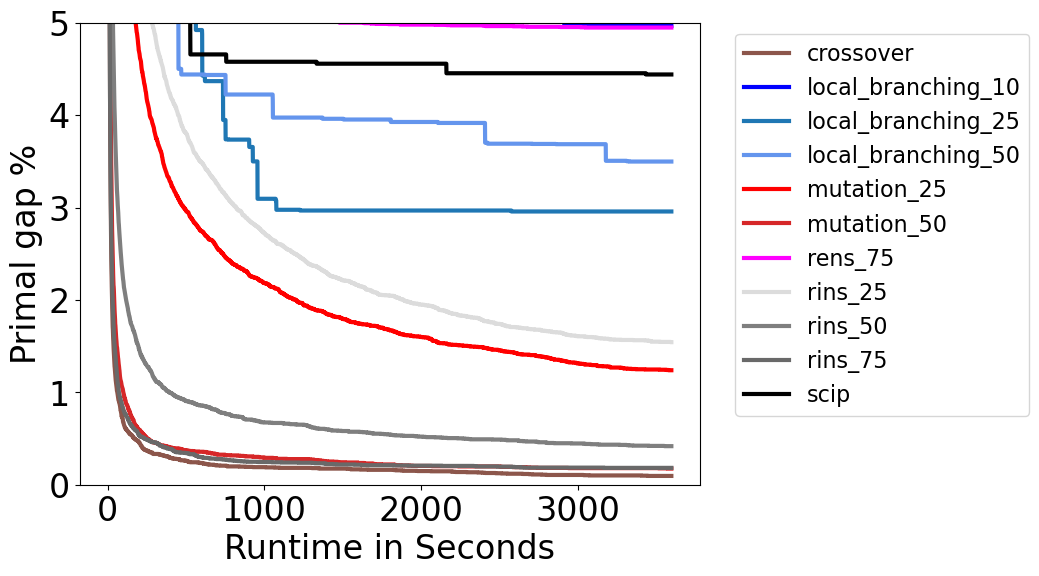}  
\end{subfigure}
\begin{subfigure}{.49\textwidth}
    \centering
    \includegraphics[width=.95\linewidth]{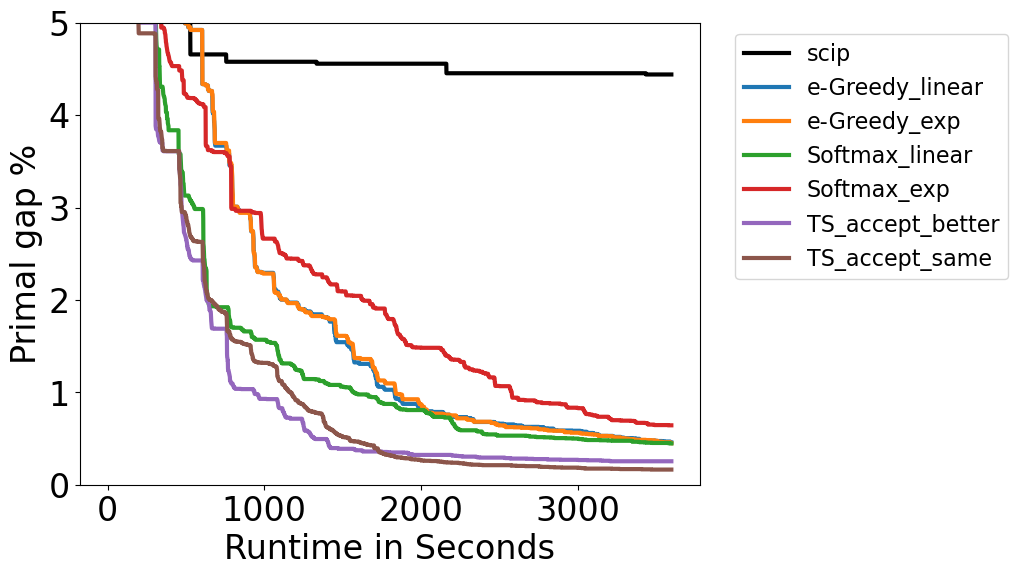}  
\end{subfigure}
\caption{The primal gap (the lower, the better) as a function of time, averaged over instances from MIS distribution}
\end{figure*}

\end{document}